\documentclass[10pt,twocolumn,letterpaper]{article}

\usepackage{cvpr}              

\usepackage{graphicx}
\usepackage{amsmath}
\usepackage{amssymb}
\usepackage{booktabs}
\usepackage{multirow}
\usepackage{times}
\usepackage{epsfig}
\usepackage{graphicx}
\usepackage{amsmath}
\usepackage{amssymb}
\usepackage{mathrsfs}
\usepackage{comment}
\usepackage{enumitem}
\usepackage{color}
\usepackage{makecell}
\usepackage{multirow}
\usepackage{float}

\newcommand\customparagraph[1]{\vspace{0.4em}\noindent\textbf{#1.}}

\usepackage[labelfont=bf,
format=plain]{caption}
\usepackage[pagebackref,breaklinks,colorlinks]{hyperref}

\usepackage[capitalize]{cleveref}
\crefname{section}{Sec.}{Secs.}
\Crefname{section}{Section}{Sections}
\Crefname{table}{Table}{Tables}
\crefname{table}{Tab.}{Tabs.}

\def\cvprPaperID{*****} 
\def\confName{CVPR}
\def\confYear{2022}

\begin{document}

\title{SuperMVS: Non-Uniform Cost Volume For High-Resolution Multi-View Stereo}

\author{Tao Zhang\\
{\tt\small atztao@gmail.com}
}
\maketitle

\begin{abstract}
  Different from most state-of-the-art~(SOTA) algorithms that use static and
  uniform sampling methods with a lot of hypothesis planes to get fine depth sampling. In this paper, we propose a
  \textbf{free-moving} hypothesis plane
  method for \textbf{dynamic and non-uniform} sampling in a wide depth range to
  build the cost volume, which not only greatly reduces the
  number of planes but also finers sampling, for both of reducing computational cost and improving
  accuracy, named \textbf{Non-Uniform Cost Volume}. We present the \textbf{SuperMVS} network
  to implement Multi-View Stereo with Non-Uniform Cost Volume. SuperMVS is a coarse-to-fine
  framework with four cascade stages. It can output higher resolution
  and accurate depth map. Our SuperMVS achieves the SOTA results with
  low memory, low runtime, and fewer planes on the DTU datasets
  and Tanks \& Temples dataset.
\end{abstract}

\section{Introduction}

3D reconstruction from multiple images is one of the important problems in
Computer Vision and Geometry. It is widely used in Autopilot, Smart City, Robot,
and AR/VR. In the 3D reconstruction, Multi-view Stereo (MVS) is an indispensable
step. It converts into a 3D model from multiple pictures and camera
parameters. Recently, some MVS methods based on Deep Learning have also achieved
good performance in 3D reconstruction.

Learning-based MVS algorithms~\cite{conf/eccv/YaoLLFQ18}~\cite{conf/cvpr/GuFZDTT20}~\cite{wang2020patchmatchnet}~\cite{conf/cvpr/YangMAL20}~\cite{conf/cvpr/ChengXZLLRS20},
build the 3D Cost Volume to estimation depth. However, the 3D Cost
Volume needs more planes subdivision the hypothesis range for higher precision
depth sampling, leading to high memory and computational cost. Such as MVSNet~\cite{conf/eccv/YaoLLFQ18} and R-MVSNet~\cite{conf/cvpr/0008LLSFQ19} 
using 256 and 512 planes for sampling the depth, their large cost volume makes optimizations take much memory and runtime. Follow-up methods like CasMVSNet~\cite{conf/cvpr/GuFZDTT20} and UCS-Net~\cite{conf/eccv/YaoLLFQ18} use cascade
architecture for coarse-to-fine the depth to reduce the number of planes to 48
and 64 planes. They can output a higher-resolution depth map with low memory and runtime. \textit{Therefore, reducing the number of planes is the key
  to reduce memory and runtime. However, how do we achieve higher accuracy with fewer planes?}

\begin{figure}[!t]
  \centering
  \setlength{\abovecaptionskip}{0.1cm}
  \setlength{\belowcaptionskip}{-0.6cm}

  {\includegraphics[width=1\linewidth]{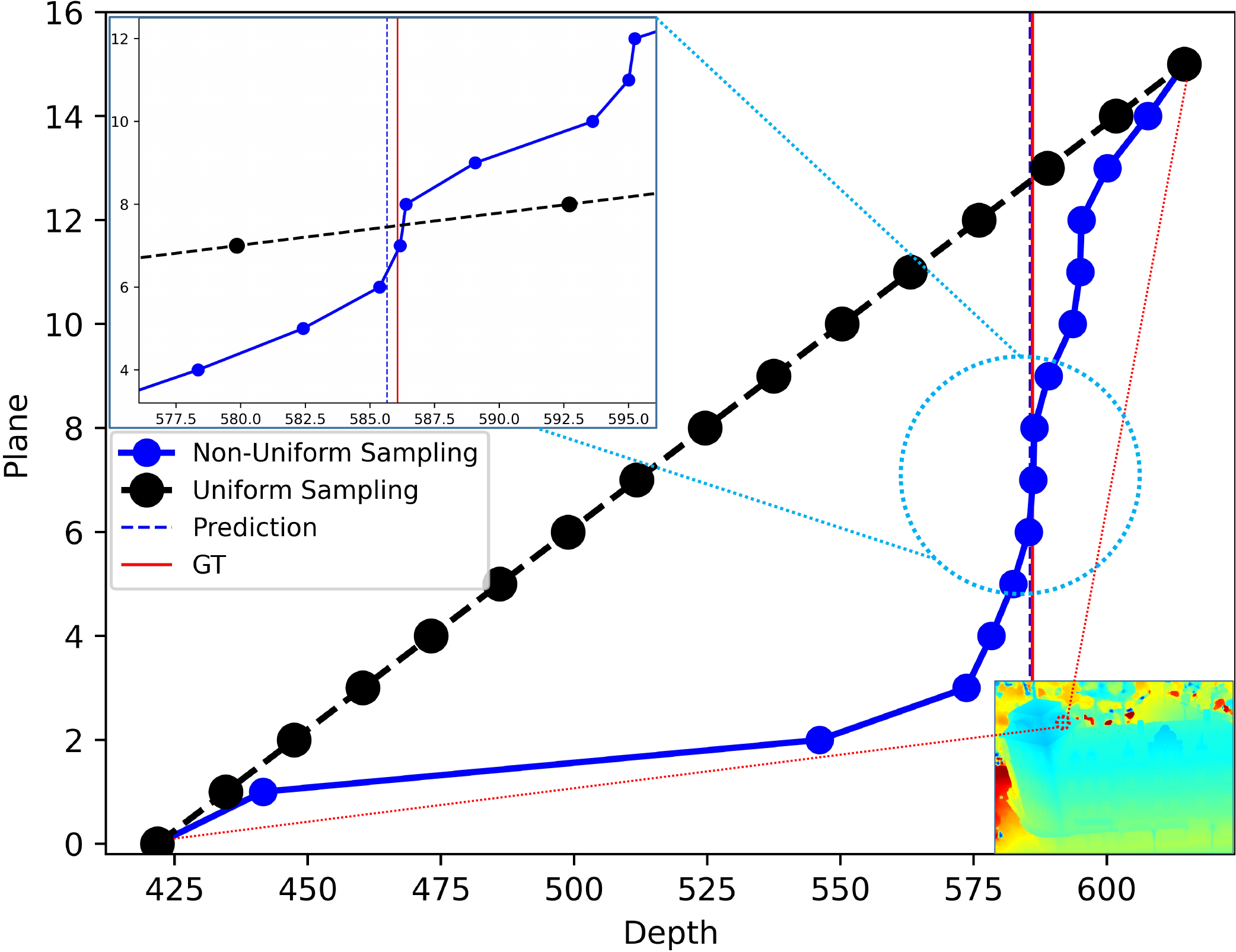}}
  \caption{The figure shows the difference between the uniform sampling method and
    non-uniform sampling method. The interval between the hypothesis planes of uniform
    sampling is equal, but the non-uniform sampling changes dynamically. Our method can make most of the planes cluster around the ground
    truth. It provides more flexible and finer interval division.}
  \label{fig:non-uniform}
\end{figure}

\begin{figure*}[htpb]
  \centering
  \setlength{\abovecaptionskip}{0.1cm}
  \setlength{\belowcaptionskip}{-0.6cm}

  {\includegraphics[width=1\linewidth]{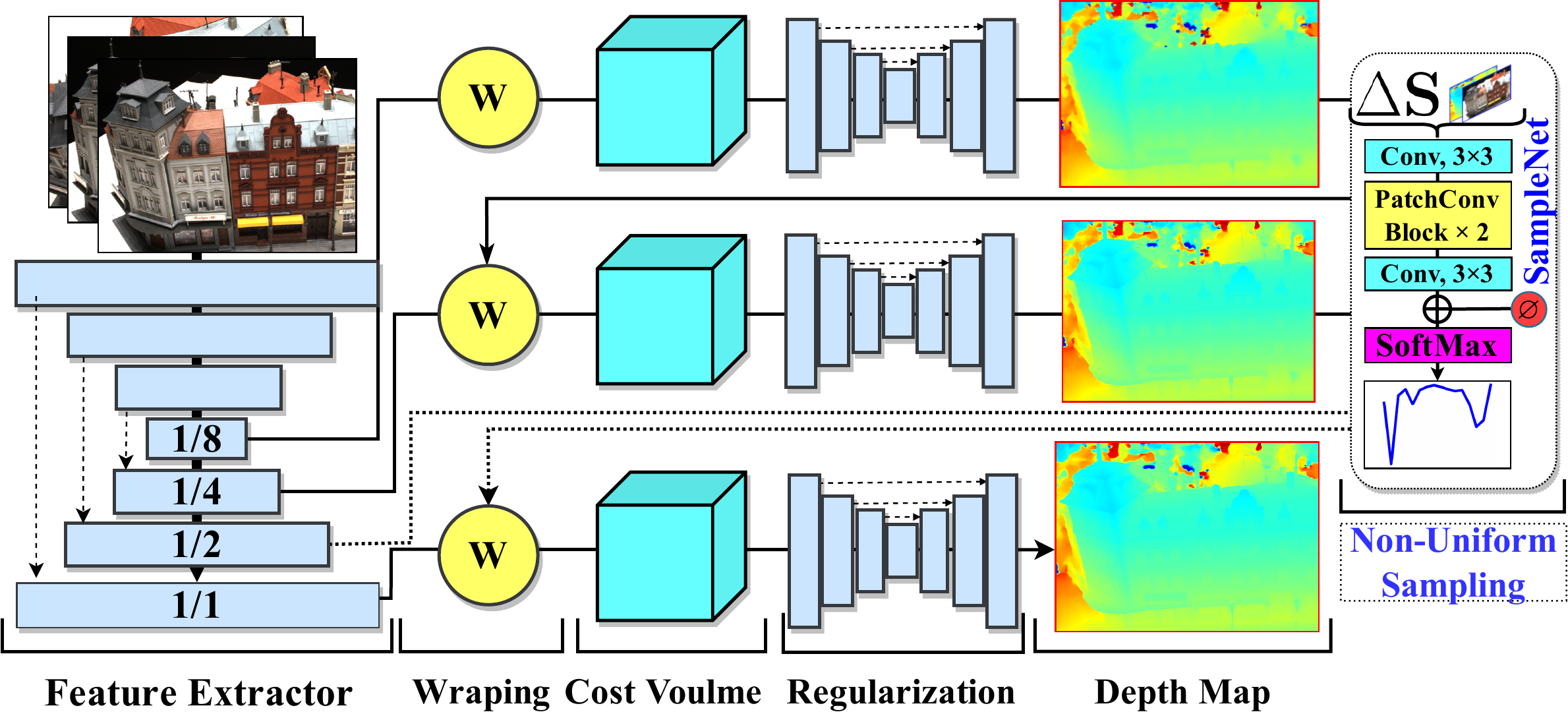}}
  \caption{SuperMVS is a coarse-to-fine architecture with four cascade stages,
    to estimate a higher accurate depth. We use the U-Net to extract four feature
    maps of the \(\{\frac{H}{8}\times \frac{W}{8}, \frac{H}{4}\times
    \frac{W}{4}, \frac{H}{2}\times \frac{W}{2},H\times W\}\) for coarse-to-fine
    the depth. All stages perform non-uniform sampling by the
    SampleNet to build the 3D cost volume, except that the first stage use the uniform
    sampling for initialize the depth.}

  \label{fig:supermvs}
\end{figure*}

In this paper, we propose a novel Non-Uniform Cost Volume method where plane moves \textit{freely} in the
depth range for \textit{dynamic and non-uniform} sampling to build the cost volume. We notice that most
algorithms use a uniform method for sampling, which require a sufficient number of planes to cover the entire depth range. The minimum depth units of uniform sampling \(\Delta d = \frac{\Delta
  R}{planes}\) depends on the number of \(planes\), larger \(planes\) requiring to the higher
fine sampling and more memory of cost volume \(h\times w \times planes\). The CNNs do not have long-range attention. Most planes
are redundant, which near the ground truth are valid. Our method is free to move most planes around the ground
truth and make nearby sampling finer to achieve higher accuracy with a fewer number of planes. In Fig.\ref{fig:non-uniform}, the interval between the plane of uniform sampling is
static and uniform. But in our method, it is dynamic and non-uniform. Our method
is more efficient than uniform sampling at the same number of planes, with a higher density of planes
distributed around the ground truth.

We present the SuperMVS
network with our Non-Uniform Cost Volume method for Multi-View Stereo, which is a coarse-to-fine
framework with four stages. To establish an efficient non-uniform sampling distribution, we
propose SampleNet in the SuperMVS. The SampleNet makes the plane
move freely in a wide depth sampling range to predict more accurate sampling. SuperMVS can output high resolution and high precision depth maps with low memory and runtime compared with other algorithms. We also achieve the
state-of-the-art result on the DTU dataset~\cite{journals/ijcv/AanaesJVTD16} and
Tanks \& Temples dataset~\cite{journals/tog/KnapitschPZK17}.




\section{Method}

Fig.\ref{fig:supermvs} shows the architecture of SuperMVS. In this section,
we describe the forward propagation process of SuperMVS. In
Sec.~\ref{feature}, we introduce the multi-scale cascade structure for feature
extraction. In Sec.~\ref{nonuniformcc}, we build non-uniform cost volume with the non-uniform
sampling. In Sec.~\ref{depthmap}, we talk about how to regularize the
non-uniform cost volume by 3D CNN to predict the depth and loss function.

The task of SuperMVS is to input \(N\) images and camera parameters, and output the depth map \( \hat{L}_{k}\). For simplicity, denote
reference image \(\mathbf{I}_{1}\), source images \(\{\mathbf{I}_i\}^{N}_{i=2}\), and camera
intrinsic and extrinsic matrices \(\{K_i,T_i\}_{i=1}^{N}\).

\subsection{Feature Extraction} \label{feature}

This module is to extract \(N\) images as feature maps \(\{F_{1,i},F_{2,i},F_{3,i},F_{4,i}\}\)
with the resolution of  \(\frac{H}{8}\times \frac{W}{8}\), \(\frac{H}{4}\times \frac{W}{4}\),  \(\frac{H}{2}\times \frac{W}{2}\) and \(H\times W\) . We build a multi-scale feature
extraction module based on U-Net~\cite{ronneberger2015convolutional}, which
includes the part of encoder and decoder. The encoder uses stride 2 convolution for
downsampling, and the decoder uses stride 2 deconvolution for upsampling. Each
convolution or deconvolution is followed by BN~\cite{ioffe2015batch} and
ReLU~\cite{agarap2018learning}. The low-scale features from the encoder are
merged by the concatenation operation. The multi-scale feature extraction
structure not only extracts more advanced features, but also reduces
computational consumption. By fusing features of different scales in the decoder
part, the redundant features can be reduced, thereby capturing more detailed
features. In our model, the channels of four stage \(\{C_{1},C_{2},C_{3},C_{4}\}\) are \(\{64,32,16,8\}\).

\subsection{Non-Uniform Cost Volume}  \label{nonuniformcc}

\customparagraph{Homography Wraping}
We use differentiable homography warping, from the feature map
\(\{F_{1,i},F_{2,i},F_{3,i},F_{4,i}\}_{i=2}^{N}\) of source views
warping to the feature map \(\{F_{1,1},F_{2,1},F_{3,1},F_{4,1}\}\) of the
reference view, to construct the cost volume. The coordinate mapping relationship
between different views is determined by homography:
\begin{equation}
  H_{i}(d)=K_{i}T_{i}T_{1}^{-1}K_{1}^{-1 }
\end{equation}
We can find the pixel coordinate \((x, y)\) and depth \(d\)
of the reference views between each source view by the homography.

\customparagraph{Non-Uniform Sampling} Assume that the depth range is \([d_{min},~d_{max}]\), the maximum
depth sampling range is \(\Delta \mathbf{R} =|d_{max}-d_{ min}|\). The number of hypothesis planes
for sampling in hypothesis range \(\Delta R_k\) at \(k\)-th stage is \(D_{k}\), \(
k\in\{1,2,3,4\}\). By the coarse-to-fine, when the depth
is more accurate, the sampling range should be reduced. Hence we define a scale 
\( R_k\) (\( 0 \leqslant R_k \leqslant 1\)) for the hypothesis range
\(\Delta R_k = R_k \Delta \mathbf{R}\).

\begin{figure}[!t]
  \centering
  \setlength{\abovecaptionskip}{0.1cm}
  \setlength{\belowcaptionskip}{-0.6cm}

  \rotatebox{0}{\includegraphics[width=1\linewidth]{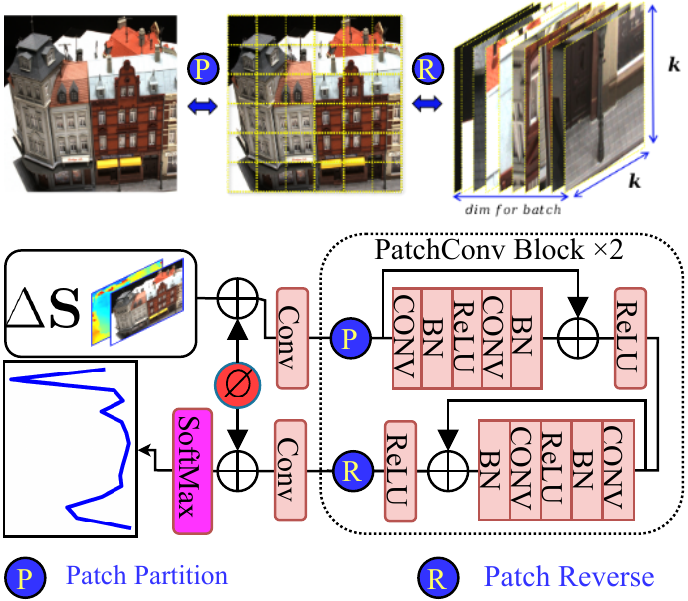}}
  \caption{SampleNet with two PatchConv blocks.}

  \label{fig:samplenet}
\end{figure}

For simplicity, we transform the sampling problem into a probability distribution problem. We purpose a SampleNet
\(\mathcal{F}\) to predict the non-uniform sampling distribution with learning: \begin{equation}
  P_{k}= \mathcal{F}(\Delta\mathbf{S}_{k-1}, \mathbf{\hat{L}}_{k-1},
  \mathbf{I}_{1}), \end{equation}
where $ \mathbf{I}_{1}$ is the reference view, $\mathbf{\hat{L}}_{k-1}$ is the
depth map of previous stage. We define the Sample Cost \(
\Delta \mathbf{S}_{k-1}\) to quantify the quality of sampling at
the previous stage and constrain the distribution of sampling at the next stage,
which is calculated from the normalized standard deviation:  \begin{equation} \begin{split}   \Delta \mathbf{S}_{k-1}  =  \frac{e^{\sqrt{(\mathbf{L}_{k-1}-\mathbf{\hat{L}}_{k-1} )^2\cdot\mathbf{P}_{k-1}}}}{\sum_{j=0}^{D_{k-1}}
      e^{\sqrt{(\mathbf{L}_{k-1,j}-\mathbf{\hat{L}}_{k-1,j}
          )^2\cdot\mathbf{P}_{k-1,j}}}}. \end{split}  \end{equation}
When the standard deviation~\(\sqrt{(\mathbf{L}_{k-1}-\mathbf{\hat{L}}_{k-1} )^2\cdot\mathbf{P}_{k-1}}\) limit to 0, it means that all
planes are distributed around the depth maps $\mathbf{\hat{L}}_{k-1}$. The depth map \(\mathbf{L}_{k-1}\) and reference view
  \( \mathbf{I}_{1}\) can also regularize the sampling distribution. In our
  experiments, SampleNet can well couple the relationship between them. There is difference in the contribution to the accuracy, $\Delta
  \mathbf{S}_{k-1}$ and  \(\mathbf{L}_{k-1}\) contributes the
  most. It will be
  discussed in Sec.~\ref{abstudy}.

From the sampling distribution $P_{k}$, we have all the
interval between the plane \( \Delta d_{k} \): \begin{equation}  \Delta d_{k} = P_k \cdot
  \Delta R_k,  \end{equation} and the sampling depth of \(j\)-th plane \(\mathbf{L}_{k,j}\) based on the
previous estimate is:
\begin{equation} \begin{aligned} \mathbf{L}_{k,j}  =\mathbf{\hat{L}}_{k-1}  - \frac{\sum^{D_k/2-1}_{j=0} \Delta d_{k,j}+ \sum^{D_k/2}_{j=0} \Delta d_{k,j}}{2} \\+ \sum_{j=0}^{j}\Delta d_{k,j} \ \ \ for\ j=0,1,\dots,D_k.  \end{aligned} \end{equation}
Specially, \( - \frac{\sum^{D_k/2-1}_{j=0} \Delta d_{k,j}+ \sum^{D_k/2}_{j=0} \Delta d_{k,j}}{2}\) makes the previous depth in
the center of all planes which is important for coarse-to-fine at next stage and
training.

\customparagraph{SampleNet With Image Patches} SampleNet (See
Fig.\ref{fig:samplenet}) uses the two PatchConv blocks, which consists of two
residual blocks~\cite{he2015residual} with inputting the image patches. Why do
we use patches as input in SampleNet?
We have tried CNN, MLP and Swin Transofmer~\cite{liu2021transformer} as the
SampleNet, but the performance of CNN and MLP is not so good as that of the Swin Transform.
The planes of CNN and MLP are distributed at both ends of the depth range. Only Swin
Transformer satisfies our distribution requirements, which is distributed around
the ground truth. We guess it is because the input of the Swin Transformer is the
image patches. The Swin
Transformer is too heavy, and we replace
its window attention modules by convolution, which greatly improves the performance. There is no good global sampling distribution for whole input,
but there is a better local distribution on local patches. In our experiments,
we set the patches size to \(8\) and the channels to \(32\) in the SampleNet. We
also add position encoding $\varnothing$ to the input and output to represent hypothesis
planes at different positions. Because the input is the patches, the sampling module
is extremely low in computation and runtime, which is discussed in Sec.~\ref{abstudy}.

\customparagraph{Non-Uniform Cost Volume}
We get the depth \(\mathbf{L}_{k,j}\) of the reference view by the non-uniform sample. Then the depth \(d\) of pixel coordinates
\((x, y)\) is \(\mathbf{L}_{k,j}(x,y)\), using the homography \(H_{i}(d)\) to
wrap the feature map \(F_{i,k}\) of the multiple source views, build the non-uniform cost
volume \(\Delta\mathbf{C}_{k,j}\).


\begin{figure*}[!ht]
  \centering
  \setlength{\abovecaptionskip}{0.1cm}
  \setlength{\belowcaptionskip}{-0.6cm}

  \rotatebox{0}{\includegraphics[width=1\linewidth]{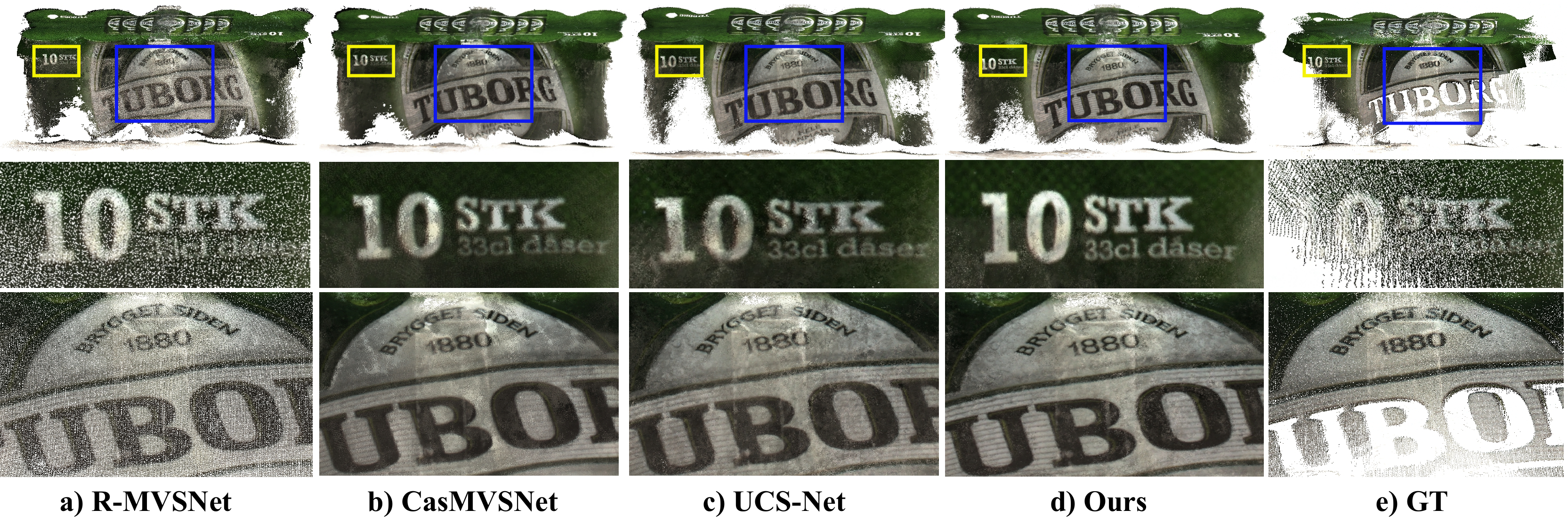}}
  \caption{Capture form the \textit{Scan12} of the DTU testing dataset. We use
    some details to reflect the density of the reconstruction and the quality of
    the point cloud. Our method's surface of the point cloud is less noisy and clearer.}

  \label{fig:vsfig}
\end{figure*}

\subsection{Depth Map and Loss Function}  \label{depthmap}

The 3D CNN is U-Net with the 3D convolution, but the skip
connections are added. In the encoder, we downsample the input resolution form
\(H \times W\) to  \(\frac{H}{8} \times \frac{W}{8}\) and the decoder
back to the input resolution by upsampling. The output layers, followed by SoftMax, output the
depth probability \(\mathbf{P}_{k,j}\).

We use the 3D CNN to regularize the 3D cost volume \(\Delta\mathbf{C}_{k,j}\)
to get the depth probability
\(\mathbf{P}_{k,j}\). The depth map \(\hat{L}_{k}\) of \(k\)-th stage is:
\begin{equation}
  \hat{L}_{k} = \sum_{j}\mathbf{L}_{k,j}\cdot\mathbf{P}_{k,j}.
\end{equation}

For supervised learning our SuperMVS, we use four smooth L1
losses~\cite{girshick2015} to build the loss function: \begin{equation}
   \mathcal{L} = \sum_{k=1}^{N=4}\lambda_{k} \cdot smooth_{L1}(\hat{L}_{k} - L^{gt}_{k}),  \end{equation}
where \(\lambda_1 = 0.25, \lambda_2 = 0.5, \lambda_3 = 1, \lambda_4 = 2\) and the \( L^{gt}_{k}\) is the ground truth of
\(k\)-th stage.


\section{Experiments}

We evaluate our method on DTU and Tanks \& Temples datasets. And Compare it with other state-of-the-art algorithms at performance, memory and runtime. We
also discuss the contribution of the Non-Uniform Cost Volume to the model.
\subsection{Datasets}
The DTU dataset~\cite{journals/ijcv/AanaesJVTD16} is a large-scale indoor MVS
dataset with 124 different scenes. It includes images with different views and lighting, camera parameters and
depth maps. It also provides the 3D point cloud data for all scenes. We use it
to train, test and validation the model. The dataset split is the same as that
in other methods
defined at~\cite{JiGZLF17}. The Tanks \& Temples dataset~\cite{journals/tog/KnapitschPZK17} consists of realistic indoor and outdoor scenes. It divided
intermediate and advanced datasets. In our work, we train the SuperMVS on DTU
training set, then test it on the DTU evaluation set and the intermediate set of Tanks
\& Temples dataset.

\subsection{Implementation}
We set the images resolution to \([H = 512,~W = 640]\) and the number of
input images to \(N=5\). The number of hypothesis plane is \(D_1
= 48\), \(D_2= 16\), \(D_3 = 8\) and \(D_4 = 8\). The scale of hypothesis range is \(R_2= 0.38\), \(R_3 = 0.16\) and \(R_4 = 0.04\). All stages use our
non-uniform sampling method to build the cost volume, except for the first stage use the uniform
sampling at \(\Delta \mathbf{R}\) for initialize the depth. We implement our
model with Pytorch~\cite{paszke2019pytorch} and train it on \(2 \times\) Nvidia
RTX 3090 GPUs with batch size of 4 per GPU. We employ the Adam~\cite{kingma2014adam}
optimizer with an initial learning rate $0.0016$ to train for $26$ epoch, where the
learning rate shcedule is divided by $2$ every two epochs starting from
$10^{th}$ epoch.

\subsection{Benchmark Performance}

\begin{table}[!h]
\begin{center}
  \footnotesize
  \resizebox{\linewidth}{!}{
\begin{tabular}{ll|cccccc}
  
  \Xhline{1pt}

  \multicolumn{2}{c|}{Method} & Acc.(\textit{mm}) & Comp.(\textit{mm}) & Overall.(\textit{mm}) & Mem.(\textit{GB}) & Runtime.(\textit{s})  & Planes. \\
  \hline\hline  
  \parbox[t]{0.5mm}{\multirow{5}{*}{\rotatebox[origin=c]{90}{Geometic}}}
  &Furu~\cite{furu} & 0.613 & 0.941 & 0.777 & - & - & -\\ 
  &Tola~\cite{tola} & 0.342 & 1.190 & 0.766 & - & - & -\\
  &Camp~\cite{camp} & 0.835 & 0.554 & 0.695 & - & - & -\\ 
  &Gipuma~\cite{galliani} & \textbf{0.283} & 0.873 & 0.578  & - & - & -\\
  &Colmap~\cite{conf/cvpr/SchonbergerF16}~\cite{conf/eccv/SchonbergerZFP16} & 0.400 & 0.664 & 0.532  & - & - & - \\
  \hline
  \parbox[t]{0.5mm}{\multirow{8.2}{*}{\rotatebox[origin=c]{90}{Learning}}}

  &SurfaceNet~\cite{JiGZLF17}  & 0.450 & 1.040 & 0.745 & - & - & -\\
  &MVSNet~\cite{conf/eccv/YaoLLFQ18} & 0.396 & 0.527 & 0.462 & \textgreater 8 & - & 256\\
  &R-MVSNet~\cite{conf/cvpr/0008LLSFQ19} & 0.383 & 0.452 & 0.417  & 6.7 & 3.72 & 512\\
  &Point-MVSNet~\cite{conf/iccv/ChenHXS19} & 0.342 & 0.411 & 0.376 & \textgreater 8 & -  & 96\\
  &Fast-MVSNet~\cite{conf/cvpr/YuG20} & 0.336 & 0.403 & 0.370 & - & - & 96\\ 
  &CasMVSNet~\cite{conf/cvpr/GuFZDTT20} & 0.325 & 0.385 & 0.355 & 6.6 & 0.76 & 48\\
  &CVP-MVSNet~\cite{conf/cvpr/YangMAL20} & 0.296 & 0.406 & 0.351 &  \textgreater 8 & - & 96 \\
  &UCS-Net~\cite{conf/cvpr/ChengXZLLRS20} & 0.338 & 0.349 & 0.344 & 6.9 & 0.84 & 64 \\
  
  \hline

  \multicolumn{2}{c|}{Ours} &0.359 & \textbf{0.293} &  \textbf{0.325}& \textbf{5.4} & \textbf{0.51} & \textbf{16}\\
  \Xhline{1pt}
  

\end{tabular}}
\end{center}
\vspace{-0.6cm}
\caption{Quantitative results on DTU's evaluation set~\cite{journals/ijcv/AanaesJVTD16} (lower is better). }
\label{tab:evaluation_dtu}
\vspace{-0.4cm}
\end{table}

\begin{table*}[!ht]
	\centering
	\begin{tabular}[width=\linewidth]{l|c|cccccccc}
		\Xhline{1pt}
		 Method &  Mean & Family & Francis & Horse & Lighthouse & M60 & Panther & Playground & Train\\ 
		\hline\hline
    MVSNet~\cite{conf/eccv/YaoLLFQ18}  & 43.48 & 55.99 & 28.55 & 25.07 & 50.79 & 53.96 & 50.86 & 47.90 & 34.69 \\
    Fast-MVSNet~\cite{conf/cvpr/YuG20}   & 47.39 & 65.18 & 39.59 & 34.98 & 47.81 & 49.16 & 46.20 & 53.27 & 42.91 \\
    Point-MVSNet~\cite{conf/iccv/ChenHXS19}   & 48.27 & 61.79 & 41.15 & 34.20 & 50.79 & 51.97 & 50.85 & 52.38 & 43.06 \\
    R-MVSNet~\cite{conf/cvpr/0008LLSFQ19}   & 48.40 & 69.96 & 46.65 & 32.59 & 42.95 & 51.88 & 48.80 & 52.00 & 42.38 \\
    PatchmatchNet~~\cite{wang2020patchmatchnet}  & 53.15 & 66.99 & 52.64 & 43.24 & 54.87 & 52.87 & 49.54 & 54.21 & 50.81 \\
    CVP-MVSNet~\cite{conf/cvpr/YangMAL20} & 54.03 & 76.50 & 47.74 & 36.34 & 55.12 & \textbf{57.28} & \textbf{54.28} & 57.43 & 47.54 \\
    UCS-Net~\cite{conf/cvpr/ChengXZLLRS20} & 54.83 & 76.09 & 53.16 & 43.03 &  54.00 &  55.60 & 51.49 & 57.38 &  47.89 \\
    CasMVSNet~\cite{conf/cvpr/GuFZDTT20} & \textbf{56.42} & \textbf{76.36} & \textbf{58.45} & \textbf{46.20} &  55.53 &  56.11 & 54.02 & \textbf{58.17} &  46.56 \\
    \hline
    Ours                   &55.38 &73.75  &54.19  &45.19  &\textbf{56.90}  &56.39  & 52.51 &55.36 & \textbf{48.78}  \\
    \Xhline{1pt}
	\end{tabular}
  \vspace{-0.2cm}
	
	\caption{Quantitative results of F-scores (higher means better) on Tanks \& Temples.}
	\label{tab:vstemplet}
  \vspace{-0.4cm}
\end{table*}

\customparagraph{Evaluation on the DTU dataset} We evaluate our method on the DTU testing set. The resolution of input
images is \([H = 1152,~W=1600]\) and the depth
range is \([425 mm,~935mm]\). The resolution of output depth map is equal to input. We reconstruct the point
cloud by fusing the depth, to calculate the \textit{accuracy},
\textit{completeness} and \textit{overall} metrics by
the MATLAB evaluate code sourced from the DTU dataset. For a fair comparison,
the GPU \textit{Mem} and \textit{Runtime} of all methods are run on Nvidia
RTX 3070 with 8GB. 

We compare our results with other state-of-the-art of learning-based methods as
the baseline in Table.\ref{tab:evaluation_dtu}. It is worth
noting that most of them use the uniform sampling method in the table. Our method outperforms other
algorithms on \textit{completeness} and \textit{overall}. We also list the number of planes set by different algorithms
when the cost volume resolution is \(\frac{H}{4} \times \frac{W}{4}  \times
planes\). We can see that
the number of planes directly affects the accuracy, memory and runtime based on
the uniform sampling method. Our model reduces computation and achieves competitive performance by using the least
number of planes 16. Compared with R-MVSNet~\cite{conf/cvpr/0008LLSFQ19},
large number of planes 512 leads to longer runtime 3.72\textit{s}, our method reduces
by 3.50\textit{s}. Compared with UCS-Net~\cite{conf/cvpr/ChengXZLLRS20},
the number of planes 64 needs 6.9\textit{GB} more memory, our method reduces the
memory by 1.5\textit{GB}. Our method also achieves significant
improvement in \textit{overall}, comparing to CasMVSNet~\cite{conf/cvpr/GuFZDTT20}
and UCS-Net, improving up 0.030\textit{mm} and 0.019\textit{mm}, respectively. In
Fig.\ref{fig:vsfig}, we compare the qualitative of point cloud with
R-MVSNet, CasMVSNet and UCS-Net. Our method has less noise and higher completeness.



\customparagraph{Evaluation on Tanks \& Temples Dataset} We also evaluate our model
on the Tanks \& Temples dataset, which is trained on the DTU dataset without any
fine-tuning. We use the image size \([H_{max} = 1024,~W_{max}=2048]\) and the number of
views $N = 10$ for input. The GPU memory and runtime spent on RTX Nvidia 3070~(8GB) are
6.5\textit{GB} and 0.88\textit{s}, respectively. We compare the F-scores of different published learning-based methods
in Table.~\ref{tab:vstemplet}. Our method outperforms most algorithms, although
belows CasMVSNet in the best average F-score, our speed and memory are superior, and we achieve the best F-scores on the Lighthouse and Train scene.

\subsection{Ablation study}  \label{abstudy}

\begin{table}[!h]
  \centering

  \resizebox{\linewidth}{!}{
    	\begin{tabular}[width=\linewidth]{ccclllcc}
      \Xhline{1pt}
      $N$  & $US$ & $NUS$ & Acc.(\textit{mm}) & Comp.(\textit{mm}) &
      Overall.(\textit{mm})  & Mem.(\textit{GB}) & Runtime.(\textit{s}) \\
      \hline\hline
      5 & \checkmark &  & 0.365   & 0.314  & 0.339 & 5.41 & 0.43 \\
      5 &  & \checkmark &  \textbf{0.357}\scriptsize{$\uparrow 0.012$} & 0.300\scriptsize{$\uparrow 0.014$} & 0.329\scriptsize{$\uparrow 0.010$} & 5.41 & 0.45 \\
      \hline

      6 &  & \checkmark &  \textbf{0.357}\scriptsize{$\uparrow 0.000$} & \textbf{0.293}\scriptsize{$\uparrow 0.007$} &  \textbf{0.325}\scriptsize{$\uparrow 0.004$} & 5.44 & 0.51  \\
      7 &  & \checkmark &  0.359\scriptsize{$\downarrow 0.002$} & 0.296\scriptsize{$\downarrow 0.003$} & 0.327\scriptsize{$\downarrow 0.002$} & 5.46 & 0.57   \\

      \Xhline{1pt}
      
    \end{tabular}}
  \caption{
    Quantitative results on DTU's evaluation
    set~~\cite{journals/ijcv/AanaesJVTD16} with the number of input views $N$.
    We training our model with uniform sampling~(\(US\)) method and non-uniform
    sampling~(\(NUS\)) method.
  }\label{tab:nus_vs_us}
\end{table}

\customparagraph{Number of Views} We evaluate the DTU's dataset with the
different number of views $N$. The results are shown in Table.~\ref{tab:nus_vs_us}.
As $N$ increases, the performance will reach saturation. When $N=6$, the performance is the best. 

\begin{table}[!h]
  \centering

  \resizebox{\linewidth}{!}{
    \begin{tabular}[width=\linewidth]{ccclll}
      \Xhline{1pt}
      $ \mathbf{I}_{1}$ & $ \mathbf{L}_{k-1}$  & $\Delta \mathbf{S}_{k-1}$ & Acc.(\textit{mm}) & Comp.(\textit{mm}) &
      Overall.(\textit{mm})  \\
      \hline\hline
      \checkmark & \checkmark & \checkmark & 0.317 & 0.237 & 0.277  \\
      \hline
      
      &  \checkmark & \checkmark & 0.322\scriptsize{$\downarrow 0.005$} & 0.238\scriptsize{$\downarrow 0.001$} & 0.280\scriptsize{$\downarrow 0.003$} \\
      &  & \checkmark & 0.359\scriptsize{$\downarrow 0.042$} & 0.276\scriptsize{$\downarrow 0.039$} & 0.318\scriptsize{$\downarrow 0.041$} \\
      &   \checkmark  &   & 0.381\scriptsize{$\downarrow 0.064$} & 0.271\scriptsize{$\downarrow 0.034$} &  0.326\scriptsize{$\downarrow 0.049$}  \\
        \checkmark &  & & 1.792\scriptsize{$\downarrow 1.475$} & 1.173\scriptsize{$\downarrow 0.936$} & 1.483\scriptsize{$\downarrow 1.206$}  \\

      \Xhline{1pt}
      
    \end{tabular}}
  \caption{
    Quantitative results on the \textit{Scan9} of DTU's evaluation
    set~~\cite{journals/ijcv/AanaesJVTD16} with the different input of SampleNet .
  }\label{tab:input_cond}
\end{table}

\customparagraph{Non-uniform Sampling Vs. Uniform Sampling} In order to fully
illustrate the superiority of our non-uniform sampling~(NUS) method, we replace it in SuperMVS with uniform
sampling~(US), train with the same training strategy as the baseline.
Table.~\ref{tab:input_cond} shows that the precision of our method~(NUS) is better
than the US, and this verifies the effectiveness of NUS. It
should be pointed out that our SuperMVS can still outperform UCS-Net and
CasMVSNet without NUS at \textit{completeness} and \textit{overall}. Comparing
memory and runtime between them, we can also find that our SampleNet occupies
almost negligible memory and runtime, and
the running time is 20\textit{ms}. 

\customparagraph{Input Contribution} Table.~\ref{tab:nus_vs_us} shows the effect
of different inputs on accuracy in SampleNet. The reference view
\( \mathbf{I}_{1}\) has the least impact on accuracy. The Sample Cost $\Delta
\mathbf{S}_{k-1}$ and the depth map $\mathbf{L}_{k-1}$ have the greatest impact
on accuracy. It shows that they are necessary to establish the correct sampling distribution.






\section{Conclusion}
This paper presents a novel dynamic and non-uniform sampling method with
fewer free-moving planes for 3D cost volume. We build the
SuperMVS for Multi-View Stereo to achieve the
purpose of low memory, low runtime and higher accuracy. We achieve the state-of-the-art performance on multiple datasets to verify our method's
effectiveness. In the future, we plan to extend the method for other
applications and combine position encoding with
2D convolution instead of 3D convolution regularization for the 3D Cost Volume.

{\small
\bibliographystyle{ieee_fullname}
\bibliography{find_lr_ref}

\begin{thebibliography}{10}\itemsep=-1pt

\bibitem{journals/ijcv/AanaesJVTD16}
Henrik Aanæs, Rasmus~Ramsbøl Jensen, George Vogiatzis, Engin Tola, and
  Anders~Bjorholm Dahl.
\newblock Large-scale data for multiple-view stereopsis.
\newblock {\em Int. J. Comput. Vis.}, 120(2):153--168, 2016.

\bibitem{agarap2018learning}
Abien~Fred Agarap.
\newblock Deep learning using rectified linear units (relu), 2018.
\newblock cite arxiv:1803.08375Comment: 7 pages, 11 figures, 9 tables.

\bibitem{camp}
Neill D.~F. Campbell, George Vogiatzis, Carlos Hernández, and Roberto Cipolla.
\newblock Using multiple hypotheses to improve depth-maps for multi-view
  stereo.
\newblock In David~A. Forsyth, Philip H.~S. Torr, and Andrew Zisserman,
  editors, {\em ECCV (1)}, volume 5302 of {\em Lecture Notes in Computer
  Science}, pages 766--779. Springer, 2008.

\bibitem{conf/iccv/ChenHXS19}
Rui Chen, Songfang Han, Jing Xu, and Hao Su.
\newblock Point-based multi-view stereo network.
\newblock In {\em ICCV}, pages 1538--1547. IEEE, 2019.

\bibitem{conf/cvpr/ChengXZLLRS20}
Shuo Cheng, Zexiang Xu, Shilin Zhu, Zhuwen Li, Li~Erran Li, Ravi Ramamoorthi,
  and Hao Su.
\newblock Deep stereo using adaptive thin volume representation with
  uncertainty awareness.
\newblock In {\em CVPR}, pages 2521--2531. IEEE, 2020.

\bibitem{furu}
Yasutaka Furukawa and Jean Ponce.
\newblock Accurate, dense, and robust multiview stereopsis.
\newblock {\em IEEE Trans. Pattern Anal. Mach. Intell.}, 32(8):1362--1376,
  2010.

\bibitem{galliani}
Silvano Galliani, Katrin Lasinger, and Konrad Schindler.
\newblock Massively parallel multiview stereopsis by surface normal diffusion.
\newblock In {\em ICCV}, pages 873--881. IEEE Computer Society, 2015.

\bibitem{girshick2015}
Ross Girshick.
\newblock Fast r-cnn, 2015.
\newblock cite arxiv:1504.08083Comment: To appear in ICCV 2015.

\bibitem{conf/cvpr/GuFZDTT20}
Xiaodong Gu, Zhiwen Fan, Siyu Zhu, Zuozhuo Dai, Feitong Tan, and Ping Tan.
\newblock Cascade cost volume for high-resolution multi-view stereo and stereo
  matching.
\newblock In {\em CVPR}, pages 2492--2501. IEEE, 2020.

\bibitem{he2015residual}
Kaiming He, Xiangyu Zhang, Shaoqing Ren, and Jian Sun.
\newblock Deep residual learning for image recognition, 2015.
\newblock cite arxiv:1512.03385Comment: Tech report.

\bibitem{ioffe2015batch}
Sergey Ioffe and Christian Szegedy.
\newblock Batch normalization: Accelerating deep network training by reducing
  internal covariate shift.
\newblock {\em arXiv preprint arxiv:1502.03167}, 2015.

\bibitem{JiGZLF17}
Mengqi Ji, Juergen Gall, Haitian Zheng, Yebin Liu, and Lu Fang.
\newblock Surfacenet: An end-to-end 3d neural network for multiview stereopsis.
\newblock In {\em ICCV}, pages 2326--2334. IEEE Computer Society, 2017.

\bibitem{kingma2014adam}
Diederik~P Kingma and Jimmy Ba.
\newblock Adam: A method for stochastic optimization.
\newblock {\em arXiv preprint arXiv:1412.6980}, 2014.

\bibitem{journals/tog/KnapitschPZK17}
Arno Knapitsch, Jaesik Park, Qian-Yi Zhou, and Vladlen Koltun.
\newblock Tanks and temples: benchmarking large-scale scene reconstruction.
\newblock {\em ACM Trans. Graph.}, 36(4):78:1--78:13, 2017.

\bibitem{liu2021transformer}
Ze Liu, Yutong Lin, Yue Cao, Han Hu, Yixuan Wei, Zheng Zhang, Stephen Lin, and
  Baining Guo.
\newblock Swin transformer: Hierarchical vision transformer using shifted
  windows, 2021.
\newblock cite arxiv:2103.14030Comment: The first 4 authors contribute equally.

\bibitem{paszke2019pytorch}
Adam Paszke, Sam Gross, Francisco Massa, Adam Lerer, James Bradbury, Gregory
  Chanan, Trevor Killeen, Zeming Lin, Natalia Gimelshein, Luca Antiga, Alban
  Desmaison, Andreas Köpf, Edward Yang, Zach DeVito, Martin Raison, Alykhan
  Tejani, Sasank Chilamkurthy, Benoit Steiner, Lu Fang, Junjie Bai, and Soumith
  Chintala.
\newblock Pytorch: An imperative style, high-performance deep learning library,
  2019.
\newblock cite arxiv:1912.01703Comment: 12 pages, 3 figures, NeurIPS 2019.

\bibitem{ronneberger2015convolutional}
Olaf Ronneberger, Philipp Fischer, and Thomas Brox.
\newblock U-net: Convolutional networks for biomedical image segmentation,
  2015.
\newblock cite arxiv:1505.04597Comment: conditionally accepted at MICCAI 2015.

\bibitem{conf/cvpr/SchonbergerF16}
Johannes~L. Schönberger and Jan-Michael Frahm.
\newblock Structure-from-motion revisited.
\newblock In {\em CVPR}, pages 4104--4113. IEEE Computer Society, 2016.

\bibitem{conf/eccv/SchonbergerZFP16}
Johannes~L. Schönberger, Enliang Zheng, Jan-Michael Frahm, and Marc Pollefeys.
\newblock Pixelwise view selection for unstructured multi-view stereo.
\newblock In Bastian Leibe, Jiri Matas, Nicu Sebe, and Max Welling, editors,
  {\em ECCV (3)}, volume 9907 of {\em Lecture Notes in Computer Science}, pages
  501--518. Springer, 2016.

\bibitem{tola}
Engin Tola, Christoph Strecha, and Pascal Fua.
\newblock Efficient large-scale multi-view stereo for ultra high-resolution
  image sets.
\newblock {\em Mach. Vis. Appl.}, 23(5):903--920, 2012.

\bibitem{wang2020patchmatchnet}
Fangjinhua Wang, Silvano Galliani, Christoph Vogel, Pablo Speciale, and Marc
  Pollefeys.
\newblock Patchmatchnet: Learned multi-view patchmatch stereo, 2021.

\bibitem{conf/cvpr/YangMAL20}
Jiayu Yang, Wei Mao, Jose~M. Alvarez, and Miaomiao Liu.
\newblock Cost volume pyramid based depth inference for multi-view stereo.
\newblock In {\em CVPR}, pages 4876--4885. IEEE, 2020.

\bibitem{conf/eccv/YaoLLFQ18}
Yao Yao, Zixin Luo, Shiwei Li, Tian Fang, and Long Quan.
\newblock Mvsnet: Depth inference for unstructured multi-view stereo.
\newblock In Vittorio Ferrari, Martial Hebert, Cristian Sminchisescu, and Yair
  Weiss, editors, {\em ECCV (8)}, volume 11212 of {\em Lecture Notes in
  Computer Science}, pages 785--801. Springer, 2018.

\bibitem{conf/cvpr/0008LLSFQ19}
Yao Yao, Zixin Luo, Shiwei Li, Tianwei Shen, Tian Fang, and Long Quan.
\newblock Recurrent mvsnet for high-resolution multi-view stereo depth
  inference.
\newblock In {\em CVPR}, pages 5525--5534. Computer Vision Foundation / IEEE,
  2019.

\bibitem{conf/cvpr/YuG20}
Zehao Yu and Shenghua Gao.
\newblock Fast-mvsnet: Sparse-to-dense multi-view stereo with learned
  propagation and gauss-newton refinement.
\newblock In {\em CVPR}, pages 1946--1955. IEEE, 2020.

\end{thebibliography}
}

\end{document}